\documentclass{article}

\usepackage[preprint]{icml2026}
\usepackage{microtype}
\usepackage{subcaption}
\usepackage{booktabs} % for professional tables

\usepackage{hyperref}
\usepackage{url}
\usepackage{amsfonts}       % blackboard math symbols
\usepackage{nicefrac}       % compact symbols for 1/2, etc.
\usepackage{xcolor} 
\usepackage{graphicx}
\usepackage{amsmath}
\usepackage{multirow}
\usepackage{enumitem}
\usepackage{pifont} % 提供带圆圈的数字符号
\usepackage{siunitx}
\usepackage{array}
\usepackage{amssymb}
\usepackage{caption} % 提供\captionof
\usepackage[export]{adjustbox} % 提供valign选项
\usepackage{tcolorbox}
\usepackage{makecell}
\usepackage{xspace,url}
\usepackage{float}
\usepackage{tabularx} % 加这个
\usepackage{colortbl}
\usepackage{wasysym}
\usepackage{wrapfig}
\usepackage{listings}
\usepackage{tikz}

\usepackage[utf8]{inputenc}
\usepackage{amsthm}

\theoremstyle{plain}

\theoremstyle{definition}

\theoremstyle{remark}

\definecolor{codebackground}{RGB}{248,249,250}
\definecolor{commentcolor}{RGB}{106,153,85}
\definecolor{keywordcolor}{RGB}{86,156,214}
\definecolor{stringcolor}{RGB}{206,145,120}
\definecolor{bordercolor}{RGB}{229,231,235}
\definecolor{titlecolor}{RGB}{75,85,99}

\definecolor{maroon}{cmyk}{0,0.1,0.01,0.01}
\definecolor{blue}{cmyk}{0.95,0.0,0.2,0.2}
\definecolor{yellow}{cmyk}{0.01,0.0,0.2,0.01}
\definecolor{lightblue}{cmyk}{0.1,0.0,0.02,0.02}
\definecolor{case_verb}{HTML}{fbde84}
\definecolor{case_adj}{HTML}{cccdff}
\definecolor{case_noun}{HTML}{bfeaf1}
\definecolor{case_ff}{HTML}{e65352}
\definecolor{case_error}{HTML}{ffff00}
\definecolor{darkgreen}{RGB}{51,181,41}
\definecolor{darkorange}{RGB}{252,135,62}
\definecolor{t_green}{HTML}{f1f2e4}

\newtcolorbox{codebox}[1][]{
    colback=codebackground,
    colframe=bordercolor,
    boxrule=1pt,
    arc=8pt,
    left=10pt,
    right=10pt,
    top=2pt,
    bottom=2pt,
    #1
}

\newcommand{\stddev}[1]{{\textcolor{gray!60}{$\pm$ #1}}}

\newcommand{\xmark}{\ding{55}}%
\newcommand{\cmark}{\ding{51}}%
\newcommand{\mycheck}{{\color{green}\cmark}}
\newcommand{\mycross}{{\color{red}\xmark}}

\newcommand{\first}[1]{\textbf{\underline{#1}}}
\newcommand{\second}[1]{\textbf{#1}}

\usepackage{mathtools}

\usepackage[capitalize,noabbrev]{cleveref}

\usepackage[textsize=tiny]{todonotes}

\newcommand{\method}{\texttt{\textbf{ICON}}\xspace}

\icmltitlerunning{ICON: Indirect Prompt Injection Defense for Agents based on Inference-Time Correction}

\begin{document}

\twocolumn[
  \icmltitle{ICON: Indirect Prompt Injection Defense for Agents \\
  based on \underline{I}nference-Time \underline{C}orrecti\underline{on}}



  \icmlsetsymbol{equal}{*}

  \begin{icmlauthorlist}
    \icmlauthor{Che Wang}{sch,yyy}
    \icmlauthor{Fuyao Zhang}{yyy}
    \icmlauthor{Jiaming Zhang}{yyy}
    \icmlauthor{ziqi Zhang}{sch}
    \icmlauthor{Yinghui Wang}{comp_m}
    \icmlauthor{Longtao Huang}{comp_a}
    \icmlauthor{Jianbo Gao}{sch}
    \icmlauthor{Zhong Chen}{sch}
    \icmlauthor{Wei Yang Bryan Lim}{yyy}
  \end{icmlauthorlist}

  \icmlaffiliation{yyy}{College of Computing and Data Science, Nanyang Technological University, Singapore}
  \icmlaffiliation{comp_a}{Alibaba}
  \icmlaffiliation{comp_m}{Ant Group}
  \icmlaffiliation{sch}{School of Computer Science, Peking University, China}

  \icmlcorrespondingauthor{Che Wang}{chewang@stu.pku.edu.cn}

  \icmlkeywords{Machine Learning, ICML}

  \vskip 0.3in
]



\printAffiliationsAndNotice{}  % no special notice (required even if empty)

\begin{abstract}
Large Language Model (LLM) agents are susceptible to Indirect Prompt Injection (IPI) attacks, where malicious instructions in retrieved content hijack the agent's execution. Existing defenses typically rely on strict filtering or refusal mechanisms, which suffer from a critical limitation: over-refusal, prematurely terminating valid agentic workflows. We propose \method, a probing-to-mitigation framework that neutralizes attacks while preserving task continuity. Our key insight is that IPI attacks leave distinct over-focusing signatures in the latent space. We introduce a \textit{Latent Space Trace Prober} to detect attacks based on high intensity scores. Subsequently, a \textit{Mitigating Rectifier} performs surgical attention steering that selectively manipulate adversarial query key dependencies while amplifying task relevant elements to restore the LLM's functional trajectory. Extensive evaluations on multiple backbones show that \method achieves a competitive 0.4\% ASR, matching commercial grade detectors, while yielding a over $50\% $ task utility gain. Furthermore, \method demonstrates robust Out of Distribution~(OOD) generalization and extends effectively to multi-modal agents, establishing a superior balance between security and efficiency.
\end{abstract}

\section{Introduction}
% \textbf{{Indirect Prompt Injection (IPI).}}
Recent advancements in Large Language Model (LLMs) Agents have revolutionized autonomous task execution, enabling sophisticated reasoning, planning, and seamless tool orchestration~\cite{Claude,openai2024gpt4technicalreport}. By integrating external data sources and diverse APIs, these agents can act upon complex user intentions within dynamic environments~(e.g., Cursor~\cite{cursor}). However, this increased agency introduces a critical security frontier: {{Indirect Prompt Injection (IPI) Attacks}}. Unlike direct jailbreaks, IPIs are embedded within retrieved content (e.g., emails, web pages). Because LLMs fundamentally conflate instructions with data, these malicious payloads can easily hijack the agent's decision-making process, forcing it to execute unauthorized actions (tool calling) while bypassing traditional safety guardrails.
\begin{figure}[t]
    \centering
    \includegraphics[width=0.48\textwidth]{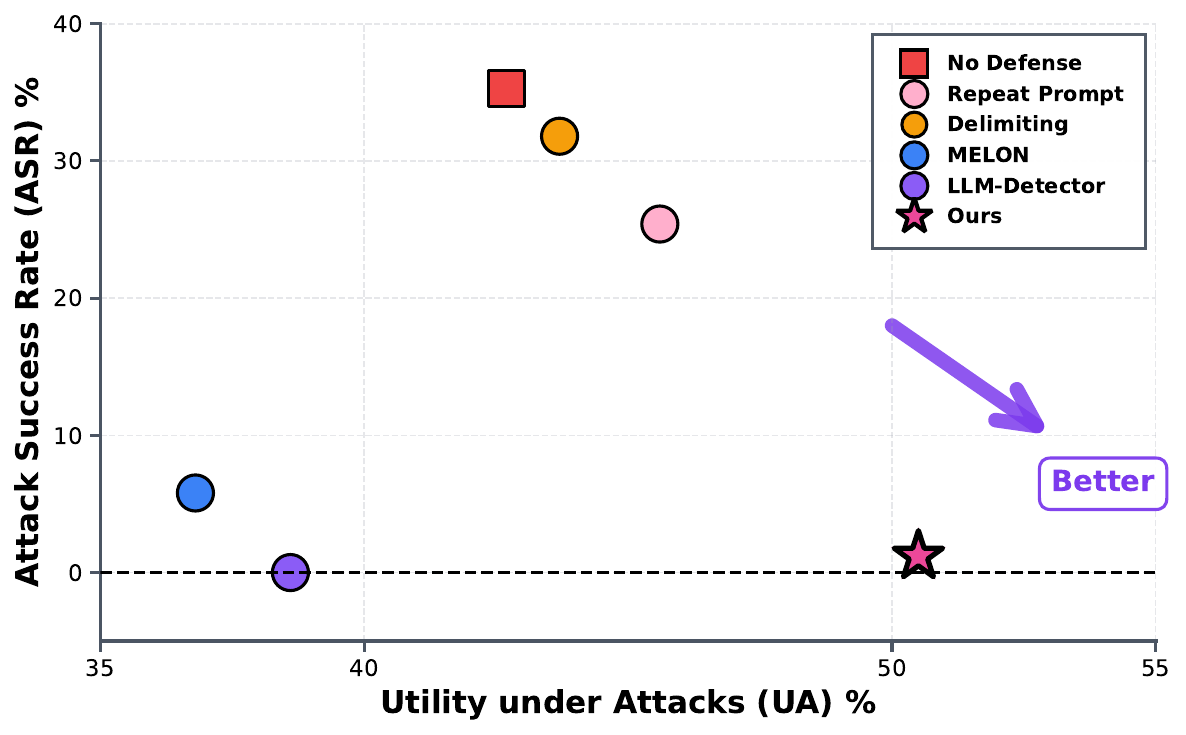} 
    \caption{Our defense, \method, presents a better security-utility trade-off (lower attack success rate, meanwhile higher utility) than other defenses against adaptive indirect prompt injection attacks.
    % \textbf{Breaking the Security-Utility Trade-off.} Comparison of averaged ASR and UA on Qwen3 across different defense methods under adaptive attacks. Our method, \method, present superior tradeoff with extremely low ASR and improving high UA.
    }
    \label{fig:trade_off}
    \vspace{-2em}
\end{figure}

% adaptive attack important
% \textbf{Adaptive IPI Attack.}
Furthermore, recent studies~\cite{zhan2024injecagent, liu2025autohijacker, zhan-etal-2025-adaptive} have highlighted the emergence of adaptive IPI attacks, which are increasingly indistinguishable from benign task contexts. By strategically exploiting the subtle semantic boundaries between instructions and data, these adversaries craft malicious payloads that bypass traditional guardrails. 
%% limitations ()

% \textbf{Limitation of Existing Defense.}
Existing solutions can not effectively defend against adaptive IPI attacks, as shown in Tab.~\ref{tab:comparison_defense}. Heuristic defenses include input templating~\cite{delimiters} and post-hoc tool filtering~\cite{zhu2025melon}. Their limitation is that they rely on surface-level patterns that are easily circumvented by diverse adversarial variants. 
While safety fine-tuning~\cite{chen2025secalign} offers stronger security, it tends to oversensitize the LLM and to excessive over-refusal, prematurely terminating complex agentic loops. Moreover, the high computational overhead of retraining makes fine-tuning impractical for agile deployment in dynamic environments. These defenses tend to sacrifice task utility for higher security.
Consequently, there is a pressing need for a plug-and-play defense mechanism that can surgically neutralize adaptive attacks without compromising the agent's functional continuity.
% Critically, these researches focus disproportionately on detection, often overlooking the imperative of maintaining holistic task utility. 

% As summarized in Tab.~\ref{tab:comparison_defense}, this evolution renders heuristic-based defenses such as input templating~\cite{delimiters} or post-hoc tool filtering~\cite{zhu2025melon} largely ineffective, as they rely on surface-level patterns that are easily circumvented by diverse adversarial variants. 
% While safety fine-tuning~\cite{chen2025secalign} offers more robust security, it often incurs a prohibitive ``safety tax'': it over-sensitizes the LLM, leading to excessive over-refusal that prematurely terminates complex agentic loops. Moreover, the high computational overhead of retraining makes fine-tuning impractical for agile deployment in dynamic environments. Critically, these researches focus disproportionately on detection, often overlooking the imperative of maintaining holistic task utility. 
% Challenges 
% Consequently, there remains a pressing need for a plug-and-play defense mechanism that can surgically neutralize adaptive attacks without compromising the agent's functional continuity.

\begin{table}[t]
\centering
\caption{Defense Comparison against Adaptive IPI Attacks.}  % 使用 \caption 而非
\label{tab:comparison_defense}
\resizebox{\linewidth}{!}{
\begin{tabular}{lcccc}
  \toprule
  Functionality & Template & Tool-Filter & Fine-tuning & \method \\
  \midrule
  Security       & \mycross & \mycross & \mycheck & \mycheck \\
  Utility        & \mycross & \mycross & \mycross & \mycheck \\
  Efficiency      & \mycheck & \mycheck & \mycross & \mycheck \\
  \bottomrule
\end{tabular}}
\vspace{-1em}
\end{table}

% \textbf{Our Defense.}
To bridge these gaps, we propose \method (\underline{I}nference-Time \underline{C}orrecti\underline{on}), a novel framework to detect and rectify compromised trajectories against adaptive IPI. Our approach is grounded in a key empirical insight: \textit{Successful IPIs that trigger a specific adversarial intent tend to induce abnormally concentrated attention on a small subset of injected tokens, compared to benign trajectories that require integrating long-horizon context.} 
We first introduce Focus Intensity Score (FIS) to quantify attention anomalies across critical layers and heads. Next, we develop the Latent Space Trace Prober (LSTP) and Mitigating Rectifier (MR). LSTP serves as a lightweight detector for high-intensity attention signatures. Upon detection, MR performs surgical attention steering rather than aborting the task. It selectively suppresses adversarial dependencies while reinforcing task-relevant features, effectively stabilizing the execution flow and restoring the agent's original intent.

%% 这一段有点多了，数字最好两个，最多三个，能够清晰得传递出delta，比现在方法强的
% \textbf{Evaluation.}
As shown in Fig.~\ref{fig:trade_off}, experiments demonstrate that $\method$ achieves superior ASR reduction and UA enhancement on out-of-distribution (OOD) benchmarks. For instance, although trained on TrojanTools~\cite{anonymous2025trojantools}, $\method$ maintains lower ASR on AgentDojo~\cite{agentdojo} than fine-tuned models (e.g., Qwen3Guard) and yields a \textbf{69\% average Utility improvement}. Furthermore, our method remains effective in multimodal LLMs,achieving \textbf{42\% average Utility recovery}. Meanwhile, $\method$ exhibits exceptional efficiency, requiring only hundreds of samples and one minute to converge. Thus, it offers a superior trade-off between security, utility, and computational cost compared to existing defenses.
Our contributions are summarized as follows:
\begin{itemize}[left=2pt, itemsep=0pt, parsep=2pt]
\vspace{-1em}
\item \textbf{Mechanism Insight:} We uncover that indirect prompt injections manifest as intrinsic ``attention collapse'' patterns, allowing for detection based on latent dynamics rather than surface-level semantics.
\item \textbf{Trajectory rectification:} We introduce a mitigating rectifier that performs targeted attention intervention on detected heads to restore benign tool-use behavior and preserve multi-step utility (achieving a $>50\%$ utility gain).
\item \textbf{Efficiency and generality:} \method trains in minutes on a small synthesized set and transfers across agents/models, including multimodal agents, achieving a strong security--utility trade-off under adaptive attacks.
\end{itemize}

\vspace{-1em}
\section{Related Works}
\label{Related_works}
\subsection{Prompt Injection Attacks \& Defense}
\textbf{Prompt injection~(PI) attacks} represents a critical vulnerability for agents. Direct prompt injections mean users attempt to bypass internal safety guardrails. Differently, indirect prompt injections~(IPI) leverage untrusted third-party data to attack. Such third-party data makes attacks much harder to detect, since adversarial instructions often appear benign within the retrieved context. 
IPI attacks typically aim to hijack an agent's tool use capabilities by embedding malicious payloads in external sources \cite{agent_safe_1, greshake2023not, zhan2025adaptive}. In this work, we focus on IPI attacks, which pose a greater threat to real-world agent scenarios nowadays.
Recent research has demonstrated how template-based prompts can mislead LLMs into executing adversaries' instructions~\cite{escape,perezignore,zhan2024injecagent}. Furthermore, studies also explored optimization-based techniques, such as using powerful LLMs or gradients to generate universal attack prompts~\cite{liu2025autohijacker,anonymous2025trojantools,zhang2024goal}. Besides, adaptive attacks have proven considerably more potent, as they can dynamically bypass static guardrails that are otherwise effective against vanilla ones.

\textbf{Prompt Injection Defense.} Recently, researchers have proposed various strategies to mitigate IPI attacks. Early efforts focused on prompt-level mitigation strategies \cite{escape, Sandwitchdefense}, which typically employ structural constraints, such as using delimiters to isolate instructions from untrusted content or reiterating system instructions within each dialogue turn to reinforce the model's adherence to the original task. Another line of research explores external detectors to intercept malicious inputs at the input stage, including Deberta detectors \cite{deberta-v3-base-prompt-injection} and training-free heuristics such as MELON \cite{zhu2025melon} and perplexity-based detection \cite{jain2023baseline_preplexity}, which are post-filter methods. However, these static approaches often struggle with the utility-security trade-off, as they may degrade model performance or introduce significant latency. Furthermore, while industry-leading models (e.g., Llama-Guard \cite{dubey2024llama3guard} and Qwen-Guard \cite{zhao2025qwen3guard}) utilize fine-tuned safety guardrails, such solutions require large-scale and high-quality safety datasets and are often computationally prohibitive for many agentic applications. In contrast, \method achieves a superior balance between security, efficiency, and utility. 

\subsection{Model Intervention Techniques}
Prior studies in model interpretability have demonstrated that internal components, particularly attention heads and layers, exhibit \textit{heterogeneous importance} across diverse tasks \cite{zhangtell, toddfunction}. For instance, induction heads \cite{crosbie2025inductionheads} specialize in in-context learning by capturing recurring patterns within input data, whereas successor heads \cite{gouldsuccessorheads} manage the incrementing of tokens in natural sequences. 
Among various interpretability driven methods \cite{chuang2024lookbackheads}, \textit{Latent Space Manipulation (LSM)} has emerged as a flexible, model-agnostic approach. Rather than modifying model parameters or architecture, LSM operates directly on internal representations such as hidden states or attention weights at inference time.

\textbf{Model Probing} is often referred to as diagnostic classification or internal state analysis \cite{hung2025attention, xia2025one}. This approach is grounded in the discovery that specific linguistic or logical tasks consistently manifest as distinct activation patterns within a model's high dimensional latent space. Afterwards, Latent Space Manipulation (LSM) can operates directly on internal representations such as hidden states or attention weights at inference time.

\textbf{Model Steering} encompasses controllable text generation techniques \cite{liang2024controllable} that guide LLMs to produce outputs with desired attributes, such as safety properties. A prominent example is LM-Steer \cite{han-etal-2024-word}, which introduces a simple yet effective mechanism to steer generation by linearly transforming the output embeddings of a frozen language model. 
Let $\mathbf{c} \in \mathbb{R}^{d}$ denote the context vector at a given decoding step. In standard decoding, the probability of generating token $v$ is calculated as:
\begin{equation}
P(v|\mathbf{c}) = \frac{\exp(\mathbf{c}^\top \mathbf{e}_v)}{\sum_{u \in V} \exp(\mathbf{c}^\top \mathbf{e}_u)},
\end{equation}
where $V$ denotes the vocabulary set and $\mathbf{e}_v \in \mathbb{R}^d$ represents the output embedding of $v$. To introduce control, LM-Steer modifies each output state using:
\begin{equation}
\mathbf{e}'_v = (\mathbb{I} + \epsilon W)\mathbf{e}_v,
\end{equation}
where $W \in \mathbb{R}^{d \times d}$ is a steering matrix, $\mathbb{I}$ is the identity matrix, and $\epsilon$ is a scalar used to adjust the steering intensity. By modifying the geometry of the output space in a controlled and interpretable manner, LM-Steer enables efficient adjustment of model behavior without additional training.

\section{Problem Formulation \& Threat Model}
\paragraph{Tool-Augmented LLM Agent.}
We adopt the formalization of tool-augmented agent $\mathcal{M}$ from prior research~\cite{yao2022react,zhu2025melon}. 
An agent is defined as a system comprising an LLM and a set of external tools, $\mathcal{F}=\{f_1, \dots, f_n\}$, for interacting with an environment. 
Given a user instruction $I_u$, the agent executes an iterative \emph{reasoning \& acting} process. 
At time step $t$, the agent maintains a trajectory state 
\begin{equation}
    \mathcal{S}_t = (I_u, \mathcal{A}_{1:t-1}, \mathcal{O}_{1:t-1}),
\end{equation}
where $\mathcal{A}_{1:t-1}$ and $\mathcal{O}_{1:t-1}$ denote the sequences of past actions and observations, respectively. 
Specifically, an action $A_i$ consists of tool calls derived from $\mathcal{F}$, while an observation $O_i$ captures execution outputs or retrieved content.

\textbf{Attacker's Capability.}
We assume a grey-box adversary with no access to the target LLM’s internal parameters or architectural details, but may infer agent context via intermediate servers. This assumption is consistent with prior work~\cite{zhu2025melon,anonymous2025trojantools}. Specifically, the attacker has three constraints: external injection, context awareness, and malicious redirection. 
External injection means the adversary can only inject a payload $\delta$ into the observation stream $\mathcal{O}_t$ via external sources (e.g., compromised MCP servers~\cite{mcp} or retrieval corpora).
Context awareness means the adversary leverages partial task knowledge~(e.g., Shopping, Payment) to ensure $\delta$ is semantically consistent with benign content, bypassing surface-level filters.
Malicious redirection means the attacker's goal is to steer the agent's policy toward an unauthorized tool call $f_{adv} \in \mathcal{F}$ while maintaining a stealthy latent trace.

\textbf{Defender's Capability.}
We define the defender as the system operator with white-box access to the local LLMs. The defender can inspect and intervene on model internals (e.g., hidden states, attention weights). Defender's goal is twofold: 1) accurately detect IPI signatures within the compromised state $\mathcal{S}_t \oplus \delta$, and 2) actively mitigate their adversarial impact during inference.
Specifically, we focus on scrutinizing the dynamics of the latent space. The defender targets the internal representations $\{\mathbf{h}^{(l)}\}_{l=1}^L$ during the critical transition from processing a compromised observation $\mathcal{O}_{1:t-1}$ to generating the subsequent action $A_t$.

\textbf{Dual Optimization Objective.} Drawing inspiration from adversarial training~\cite{zhang2025anyattack, zhang2025dualtap}, we formalize the defense against IPI attacks as a decoupled optimization problem consisting of two objectives: the attacker's goal and defender's goal. 

\textbf{Attacker's Goal.}
To generate the most challenging samples for the defender and maximize stealthiness, the adversary seeks a context-aware instruction $\delta$ that minimizes the discrepancy between the compromised and benign reasoning flows. The attack should also successfully trigger the malicious intent $f_{adv}$. The objective is defined as:
\begin{equation}
\min_{\delta} \mathcal{D}(S_t \oplus \delta, S_t) \quad \text{s.t.} \quad \mathcal{M}(S_t \oplus \delta) = f_{adv},
\label{attack_target}
\end{equation}
where $S_t$ is the benign trajectory, $\oplus$ denotes the injection operation, and $\mathcal{D}$ is a discrepancy function (i.e., LLM-as-Optimizer) analyze and optimize the stealthiness of $\delta$. 

\textbf{Defender's Goal.}
Given the adaptive trajectory from the attacker, the defender aims to jointly minimize the detection error and maximize the task utility. We formalize this as a dual-objective optimization of the prober parameters $\phi$ and the steering hyperparameter $\gamma$ and $\tau$ of the rectification operator $\mathcal{R}$:
\begin{equation}
\min_{\phi, \gamma} \Big[ \mathcal{L}(h_{\phi}, y) - \log P_{\mathcal{M}}\big(y \mid \mathcal{R}(S_t \oplus \delta; \gamma; \tau)\big) \Big],
\end{equation}
where $\mathcal{L}$ is the detection loss of the prober against ground truth $y$. The second term maximizes the log-probability of the correct task output $y$ under the rectification operator $\mathcal{R}$.
Here, $\mathcal{R}(\cdot; \gamma; \tau)$ denotes the inference-time attention steering controlled by $\gamma$ and $\tau$, which restores the agent's focus without updating model weights.

\section{Methodology}

\begin{figure*}[t]
    \centering
    \includegraphics[width=0.95\linewidth]{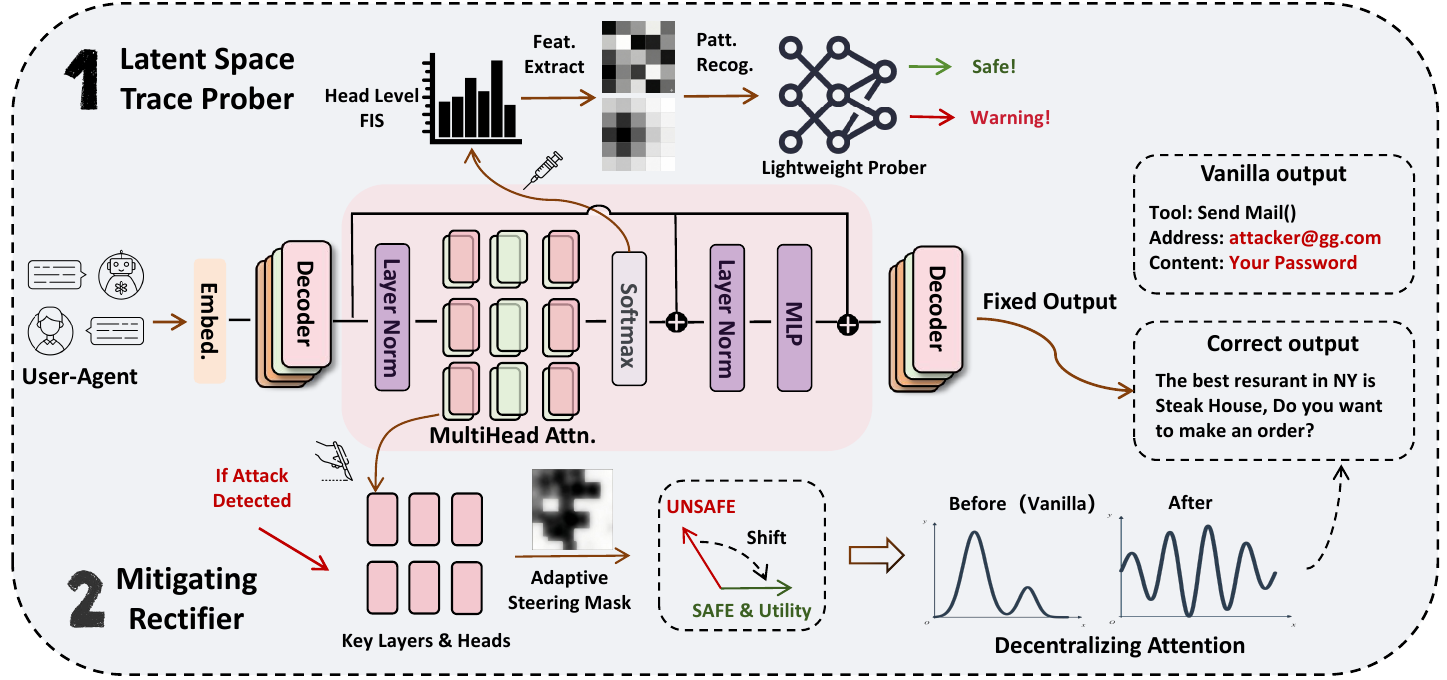}
    \caption{\textbf{The overall architecture of \method.} The framework comprises two core modules: 
    (1) The \textbf{Latent Space Trace Prober} for identifying intrinsic anomalies in the model's internal representations, and (2) The \textbf{Mitigating Rectifier} for neutralizing threats and restoring functional integrity. Together, these mechanisms enable robust detection and effective rectification of indirect prompt injections.}
    \label{fig:framework}
\end{figure*}

\subsection{Motivation}
\method is inspired by the contrast between an agent's \textit{holistic context integration} and the \textit{localized focus} inherent in IPI attacks. While benign reasoning relies on long-horizon dependencies within the trajectory $\mathcal{S}_t$ for planning consistency, a successful injection necessitates a ``forced focus'' on the adversarial segment to trigger specific target tools $f_{adv}$. This structural anomaly in the attention mechanism makes the model prioritizes the malicious payload over global task logic, thus creates a distinct latent trace that allows \method to differentiate genuine reasoning from adversarial redirection.

\subsection{Overview}
The design of $\method$ is driven by \textbf{two critical challenges} in safeguarding LLM agents. First, adversarial diversity: current heuristic-based defenses (e.g., tool filtering~\cite{zhu2025melon} or static prompts~\cite{delimiters}) are easily bypassed by adaptive attacks that seamlessly blend malicious intent into the agent's context. Second, execution continuity: conventional safety mechanisms predominantly focus on binary refusal, which prematurely terminates the agent's multi-step reasoning loop and severely degrades task utility.

To address these, we propose a pipeline that first generates adaptive attacks (Sec.~\ref{adaptive_attack}) to delineate and expand the agent's safety boundaries against benign contexts~(Offline) for training defense modules. Then, it is followed by an online framework for real-time detection and utility recovery (Sec.~\ref{defend}). The defense mechanism, illustrated in Figure~\ref{fig:framework}, follows a two-step paradigm to ensure both \textit{Robust Security} and \textit{Task Utility}. The Latent Space Trace Prober (LSTP) serves as a real-time gatekeeper, scrutinizing the agent's latent space for anomalous signals. Upon detection, the Mitigating Rectifier (MR) adaptively steers attention weights. By selectively suppressing adversarial query-key dependencies and redistributing focus toward the benign task context, the MR restores functional integrity, ensuring the agent completes its trajectory without execution loop termination.

\subsection{Adaptive Attack Data Synthesis}
\label{adaptive_attack}
To provide high-quality training data for the prober and calibrate the rectifier, we develop an offline data synthesis module. The primary goal of this module is to generate boundary-case IPI samples that satisfy the optimization objective in Eq.\ref{attack_target}. Unlike static datasets, our approach simulates a context-aware adversary to ensure the synthesized attacks are semantically seamless and latent-stealthy. We employ an {LLM-as-Optimizer} framework to iteratively solve the attacker's objective. 

This process focuses on two dimensions of the attack space:
\textbf{(a).~Contextual Tool Alignment}: Instead of random injection, the optimizer selects the most plausible malicious tool $f_{adv} \in \mathcal{F}$ that aligns with the current trajectory $\mathcal{S}_t$. This ensures that the final attack samples do not trigger surface-level semantic filters due to logical inconsistency.
\textbf{(b).~Iterative Stealthiness Optimization}: Starting from a random initialization, the optimizer refines the instruction $\delta$ to minimize the discrepancy $\mathcal{D}$. This is achieved through a closed-loop feedback mechanism: the optimizer analyzes the agent's prior failure or success attempts and adaptively updates the payload towards the benign samples in the latent space while still fulfilling the success constraint $\mathcal{M}(S_t \oplus \delta) = f_{adv}$.

This synthesis process is not intended to study the attack itself, but rather to delineate the adaptive-benign boundary. By generating a diverse set of stealthy instructions across various agentic tasks, we construct a robust training distribution. This allows the subsequent LSTP and MR to learn the fundamental over-focusing signal of IPI attacks rather than over-fitting to specific keyword patterns.

\subsection{Probing and Mitigating Techniques}
\label{defend}
In this section, we introduce three synergistic components: the Focus Intensity Score~(FIS), the Latent Space Trace Prober~(LSTP), and the Mitigating Rectifier~(MR). As the diagnostic foundation, the FIS quantifies attention entropy to identify anomalous focusing patterns, providing the critical head-selection signals required by both the LSTP for precise detection and the MR for surgical intervention. By integrating these mechanisms, $\method$ effectively transitions from identifying latent adversarial traces to performing real-time attention steering, ensuring robust security without compromising the agent's functional utility.

\subsection{Focus Intensity Score (FIS)}
To quantify the specialization of attention heads during the generation process, we introduce the {Focus Intensity Score}. Our key insight is that \textit{attention heads with lower entropy tend to capture specific, task-critical semantic patterns, which are often exploited or disrupted during IPI attacks.}

Unlike previous methods that only consider the last token of input or start token of generation~\cite{hung2025attention}, we analyze the entire generated sequence $Y$ to evaluate its global dependency on the context $\mathcal{S}_t$. Let $L$ and $H$ denote the total number of layers and the number of heads per layer, respectively. For a generated sequence of length $N_Y$, the attention weight matrix of the $h$-th head in layer $l$ is denoted as $\mathbf{A}_{l,h} \in \mathbb{R}^{N_Y \times N}$, where $N$ is the sequence length of $\mathcal{S}_t$(current input). Each element $a_{i,j}$ represents the attention weight from the $i$-th generated token (query) to the $j$-th context token (key).

We first compute the generation-normalized token entropy for each query $i$ to align the focus intensity with the generation scale:
\begin{equation}
    E(l, h, i) = - \frac{1}{\log N} \sum_{j=1}^{N} a_{i,j} \log(a_{i,j} + \epsilon),
\end{equation}
where $\epsilon$ is a small constant for numerical stability. Denote {FIS} for a specific head as $S_{l,h}$. FIS can be defined as the complement of the average entropy across the entire generated sequence. And the aggregate score for layer $l$ is the mean FIS of its constituent heads:
\begin{equation}
    S_{l,h} = 1 - \frac{1}{N_Y} \sum_{i=1}^{N_Y} E(l, h, i), \quad S_l = \frac{1}{H} \sum_{h=1}^{H} S_{l,h}
\end{equation}

\subsection{Latent Space Trace Prober (LSTP)}
The Latent Space Trace Prober (LSTP) is a lightweight detector for real-time identification of IPI attacks during inference. To ensure robust {Out-of-Distribution (OOD)} generalization across diverse scenarios, the prober should capture critical adversarial patterns in the latent space rather than memorize specific parameter values. To distill consistent signatures from raw attention weights, we develop a multi-stage transformation that maps variable-length sequences into a fixed-dimensional feature space. This process effectively isolates the ``attention sink'' dynamics characteristic of autoregressive transformers, allowing the prober to detect subtle deviations induced by adversarial instructions.

\textbf{Temporal Feature Aggregation.} As the number of generated tokens $N_Y$ varies per sample, we apply a temporal aggregation function $\mathcal{G}$ to compress the entropy sequence $\{E_{l,h,i}\}_{i=1}^{N_Y}$ into a fixed-length descriptor. We select three significant features to characterize the focus dynamics of each head $h$:
\begin{itemize}[left=5pt, itemsep=2pt, parsep=0pt]
    \item {Minimum Entropy}: Identifies the most extreme collapse of attention, often signifying an intensive response to an adversarial trigger.
    \item {Mean Entropy}: Reflects the overall concentration of the head throughout the generation process.
    \item {Standard Deviation}: Measures the volatility and stability of the attention focus across time steps.
\end{itemize}
Consequently, the feature vector for head $h$ in layer $l$ is defined as $\mathbf{v}_{l,h} \in \mathbb{R}^3$:
\begin{equation}
\begin{aligned}
    \mathbf{v}_{l,h} = \Big[ & \min_{i} E_{l,h,i}, \frac{1}{N_Y} \sum_{i=1}^{N_Y} E_{l,h,i}, \\
    & \sqrt{\frac{1}{N_Y} \sum_{i=1}^{N_Y} (E_{l,h,i} - \bar{E}_{l,h})^2} \Big] \in \mathbb{R}^3
\end{aligned}
\end{equation}

\textbf{Feature Fusion and Prober Architecture.} To construct a comprehensive representation, we select a subset of $K$ attack sensitive layers $\mathcal{L}^* \subseteq \{1, \dots, L\}$, based on \textbf{FIS} of each layer. The global feature vector $\mathbf{z}$ is formed by concatenating head-wise features across all selected layers:
\begin{equation}
    \mathbf{z} = \text{Concat}\left( \{ \mathbf{v}_{l,h} \mid \forall h \in \{1, \dots, H\}, \forall l \in \mathcal{L}^* \} \right)
\end{equation}
For example, if a model with $H=32$ heads and $K=4$ layers, this yields a representation $\mathbf{z} \in \mathbb{R}^{3 \times 32 \times 4} = \mathbb{R}^{384}$. 

To capture local dependencies and sequential anomalies in attention shifts, we employ a hybrid architecture comprising a 1D-Convolutional Neural Network (CNN) and a Multi-Layer Perceptron (MLP). The CNN layers facilitate the extraction of inter-head and inter-layer correlations. Furthermore, to ensure invariance to response duration, we utilize an \textit{Adaptive Global Maximum Pooling (GMP)} layer. By extracting the most salient activations across the generation timeline, the prober can effectively identify both short, aggressive injections and long, subtle adversarial redirections:
\begin{equation}
    \mathbf{z}_{final} = \text{AdaptiveMaxPool1d}(\mathbf{H}^{(final)}) \in \mathbb{R}^{d_{feat}}
\end{equation}
where $\mathbf{H}^{(final)}$ denotes the high-level feature maps before the output layer. This architectural design ensures that the prober maintains high computational efficiency while preserving superior sensitivity to intrinsic latent traces, thereby facilitating robust generalization even to OOD datasets.

\subsection{Mitigating Rectifier (MR)}
Prior studies~\cite{arditi2024refusal, wu2025automating} primarily rely on refusal vectors to thwart attacks. However, such interventions often induce prematurely termination of the execution loop, a fatal disruption for multi-step autonomous agents. To address this, this module performs a surgical latent rectification rather than a binary rejection. 

Upon detection, the MR executes attention functionality restoration. By suppressing anomalous query-key dependencies while implicitly amplifying task-relevant context, it steers the model's latent representation back toward the benign functional manifold. This ensures the agent neutralizes malicious instructions without compromising the semantic continuity of its original plan.

\textbf{Dual-Factor Attention Intervention.} To ensure precision, we introduces a dual-hyperparameter mechanism that decouples the \textbf{scope} ($\tau$) and \textbf{intensity} ($\gamma$) of the intervention:

\begin{itemize}[leftmargin=15pt, itemsep=1pt]
    \item \textbf{Steering Scope ($\tau$):} For each identified anomalous head $h \in \mathcal{H}_{adv}$ (based on FIS), we define a suppression threshold $\theta_{l,h}$ governed by the top-$k$ ratio $\tau$. Specifically, $\theta_{l,h}$ is determined as the $(1-\tau)$-th percentile of the attention weights in head $h$, defining the \textit{saliency budget} for intervention. The steering mask $M_{l,h}$ is formally defined as:
    \begin{equation}
        M_{l,h}(i,j) = \mathbb{I} \big( a_{l,h}(i,j) \geq \theta_{l,h} \big),
    \end{equation}
    where $a_{l,h}(i,j)$ is the raw attention weight and $\mathbb{I}(\cdot)$ denotes the indicator function. 

    \item \textbf{Steering Intensity ($\gamma$):} We apply a \textit{Contrastive Steering Operation} to the attention matrix using the coefficient $\gamma$:
    \begin{equation}
        \tilde{A}_{l,h} = A_{l,h} \odot [1 + M_{l,h} \cdot (\gamma - 1)].
    \end{equation}
\end{itemize}

In this formulation, $\gamma < 1$ functions as an intensive suppression coefficient. This mechanism is conducted before multi-head attention softmax mechanism that exploits the Softmax redistribution effect: by diminishing anomalous attention peaks, the subsequent normalization naturally reallocates focus toward the benign task context (where $M_{l,h}=0$). This selective re-weighting effectively performs latent-level signal-to-noise enhancement, restoring the agent's functionality without damaging the global semantic structure.

\begin{table*}[t]
\centering
\small
\setlength{\tabcolsep}{2.5pt}
\renewcommand{\arraystretch}{1.2}
\caption{Evaluation of \method against IPI attacks on various agentic LLMs.}
\resizebox{0.95\textwidth}{!}{%
\begin{tabular}{l|l|l|cccccccccccc}
\toprule
\multirow{2}{*}{\textbf{FMs}} & \multirow{2}{*}{\textbf{BU}} & \multirow{2}{*}{\textbf{Attack}}
& \multicolumn{2}{c}{\textbf{No Defense}} & \multicolumn{2}{c}{\textbf{Repeat Prompt}} 
& \multicolumn{2}{c}{\textbf{Delimiting}} & \multicolumn{2}{c}{\textbf{MELON}} 
& \multicolumn{2}{c}{\textbf{LLM-Detector}} & \multicolumn{2}{c}{\textbf{ICON~(Ours)}}\\
\cmidrule(lr){4-5} \cmidrule(lr){6-7} \cmidrule(lr){8-9} \cmidrule(lr){10-11} \cmidrule(lr){12-13} \cmidrule(lr){14-15}
& & & ASR$\downarrow$ & UA$\uparrow$ & ASR$\downarrow$ & UA$\uparrow$ & ASR$\downarrow$ & UA$\uparrow$ & ASR$\downarrow$ & UA$\uparrow$ & ASR$\downarrow$ & UA$\uparrow$ & ASR$\downarrow$ & UA$\uparrow$ \\
\midrule
\multirow{4}{*}{\textsc{Qwen-3-8B}} 
& \multirow{4}{*}{69.8} & Ignore Instruction & 25.0 & 49.7 & 11.7 & 54.2 & 15.3 & 52.9 & 1.3 & 48.7 & 0.0 & 47.2 & 0.0 & 58.2 \\
& & Combined Attacks & 30.3 & 44.2 & 13.2 & 53.1 & 18.4 & 50.2 & 2.4 & 45.5 & 0.0 & 43.8 & 0.0 & 52.7 \\
& & TrojanTools & 42.7 & 35.3 & 25.4 & 45.6 & 31.8 & 43.7 & 5.8 & 36.8 & 0.6 & 38.6 & 1.2 & 50.5 \\
\cmidrule(lr){3-15}
& & \textbf{Avg} & 32.7 & 43.1 & 16.8 & \second{51.0} & 21.8 & 48.9 & 3.2 & 43.7 & \first{0.2} & 43.2 & \second{0.4} & \first{53.8} \\
\midrule
\multirow{4}{*}{\textsc{LLaMA-3.1-8B}} 
& \multirow{4}{*}{34.5} & Ignore Instruction & 23.4 & 25.7 & 14.6 & 30.1 & 16.8 & 28.1 & 0.6 & 26.1 & 0.0 & 25.2 & 0.0 & 32.6 \\
& & Combined Attacks & 36.7 & 22.2 & 20.3 & 28.6 & 22.4 & 25.8 & 0.8 & 21.9 & 0.0 & 22.5 & 0.0 & 29.3 \\
& & TrojanTools & 47.1 & 14.3 & 28.9 & 25.5 & 30.5 & 24.6 & 3.5 & 14.8 & 0.0 & 15.1 & 0.8 & 27.1 \\
\cmidrule(lr){3-15}
& & \textbf{Avg} & 35.7 & 20.7 & 21.3 & \second{28.1} & 23.2 & 26.2 & 1.6 & 20.9 & \first{0.0} & 20.9 & \second{0.3} & \first{29.7} \\
\midrule
\multirow{4}{*}{\textsc{Mistral-8B}} 
& \multirow{4}{*}{20.1} & Ignore Instruction & 16.3 & 12.5 & 10.3 & 17.8 & 12.1 & 16.5 & 0.2 & 12.1 & 0.0 & 11.4 & 0.0 & 17.6 \\
& & Combined Attacks & 33.6 & 11.1 & 19.8 & 15.6 & 23.6 & 14.1 & 0.5 & 11.5 & 0.0 & 10.9 & 0.0 & 16.1 \\
& & TrojanTools & 42.9 & 8.5 & 23.2 & 12.3 & 27.9 & 10.2 & 2.6 & 9.3 & 0.0 & 9.7 & 0.9 & 13.9 \\
\cmidrule(lr){3-15}
& & \textbf{Avg} & 30.9 & 10.7 & 17.8 & \second{15.2} & 21.2 & 13.6 & 1.1 & 11.0 & \first{0.0} & 10.7 & \second{0.3} & \first{15.9} \\
\bottomrule
\end{tabular}}
\label{tab:defense_results_nlp}
\begin{flushleft}
\scriptsize
\textbf{Notes:} 
\textit{ASR} = Attack Success Rate (\%, lower is better for defender).
\textit{UA} = Utility under Attack (\%, higher is better for defender).
\textit{LLM-Detector} = Qwen3Guard.
\first{Bold} indicates best result.
\second{Bold} indicates second-best UA.
\second{\method maximizes UA with negligible impact on ASR.}
\end{flushleft}
\vspace{-1em}
\end{table*}

\section{Experiments}
\label{experiments}
\subsection{Experimental Setup}
We evaluate the effectiveness of \method by integrating it into the {ReAct}~\cite{yao2022react} framework, with a specific focus on the mitigation of IPI attacks. Our experiments are conducted across three representative foundation LLMs: \textsc{Qwen}~\citep{yang2025qwen3}, \textsc{LLaMA}~\cite{touvron2023llama}, and \textsc{Mistral}~\cite{mistral}. For Visual Large Models, we utilize \textsc{Qwen-VL}, \textsc{InternVL}~\cite{chen2024internvl}, and \textsc{MiniCPM}~\cite{yu2025minicpm}. These models were selected for their diverse architectures, ranging from reasoning-optimized to general-purpose designs, and their robust native support for tool-calling mechanisms.

\textbf{Datasets.} We utilize two widely used datasets, \emph{InjectAgent}~\cite{zhan2024injecagent} and \emph{AgentDojo}~\cite{agentdojo} to evaluate the effectiveness of \method. For training, we adopt TrojanTools~\cite{anonymous2025trojantools}, which provides diverse tool selections in order to demonstrate the out-of-distribution capability of our \method. For visual tasks, we utilize Visual Prompt Injection Benchmarks~\cite{wan2024CyberSecEval} provided by Meta.

\textbf{Baselines.} 
Following recent literature~\citep{zhu2025melon, bhagwatkarindirect}, we benchmark \method against three defense mechanisms: template-based, filter-based, and fine-tuning-based.
Template-based defenses include {Repeat Prompt}~\cite{agentdojo}, which reiterates system instructions, and {Delimiting}~\cite{delimiters}, which uses special tokens to isolate untrusted content.
For filtering-based defense, we use {MELON}~\cite{zhu2025melon}, a state-of-the-art tool-filtering framework. 
For fine-tuning-based defense, we utilize {Qwen3Guard}~\cite{zhao2025qwen3guard} and Gemini~\cite{team2023gemini}, the most advanced (M)LLM currently available, to establish a performance upper bound for intrinsic model robustness. 
%We provided the attack methods selection details in Appendix.\ref{Attack_appendix}.
% 
% For the attacks, we employ three representative IPI attack methods: 1) {Ignore Instruction}~\cite{schulhoff2023ignore}, a traditional prompt-based hijacking technique; 2) {Combined Attack}~\cite{zhan2024injecagent}, an attack integrated method specifically designed for agentic tool-use scenarios; and 3) {TrojanTools}~\cite{anonymous2025trojantools}, a state-of-the-art adaptive attack that mimics legitimate conversation flows.

\begin{table}[t]
\centering
\small
\setlength{\tabcolsep}{6pt}
\renewcommand{\arraystretch}{1.3}
\caption{OOD Performance of \method trained on TrojanTools.}
\resizebox{\columnwidth}{!}{%
\begin{tabular}{ll cccc}
\toprule
\multirow{2}{*}{\textbf{Model}} & \multirow{2}{*}{\textbf{Metric}} & \multicolumn{2}{c}{\textbf{InjectAgent}} & \multicolumn{2}{c}{\textbf{AgentDojo}} \\
\cmidrule(lr){3-4} \cmidrule(lr){5-6}
& & \textbf{No Def.} & \textbf{ICON (Ours)} & \textbf{No Def.} & \textbf{ICON (Ours)} \\
\midrule
\multirow{2}{*}{\textsc{Qwen-3-8B}} 
& ADR & 0\% & \textbf{80.1\%} \stddev{1.9} & 0\% & \textbf{98.0\%} \stddev{1.2} \\
& URR & 0\% & \textbf{62.3\%} \stddev{2.3} & 0\% & \textbf{69.6\%} \stddev{2.7} \\
\midrule
\multirow{2}{*}{\textsc{LLaMA-3.1-8B}} 
& ADR & 0\% & \textbf{85.4\%} \stddev{1.3} & 0\% & \textbf{97.3\%} \stddev{1.8} \\
& URR & 0\% & \textbf{47.6\%} \stddev{3.1} & 0\% & \textbf{55.9\%} \stddev{3.3} \\
\bottomrule
\end{tabular}}
\begin{flushleft}
\scriptsize
\textbf{Notes:} \textit{ADR} = Attack Detection Rate. \textit{URR} = Utility Recover Rate.
Values are reported with standard deviations computed over 5 cross validation.
\end{flushleft}
\label{tab:detection_defense_rate}
\end{table}

\begin{table}[t]
\centering
\caption{Efficiency comparison of different methods.}
\label{tab:efficiency}
\resizebox{\columnwidth}{!}{%
\begin{tabular}{lccc}
\toprule
\textbf{Method} & \textbf{Training Cost} & \textbf{Dataset Size} & \textbf{Parameters} \\
\midrule
Qwen3Guard & Over 10 hours & 1.19M & 8B \\
MELON & 0 & 0 & 0 \\
Repeat Prompt & 0 & 0 & 0 \\
\midrule
\textbf{ICON (Ours)} & \textbf{Less than 2 mins} & \textbf{255} & \textbf{31553} \\
\bottomrule
\end{tabular}}
\end{table}

\subsection{Main Results and Analysis}
Table~\ref{tab:defense_results_nlp} summarizes the performance of \method across three backbone LLMs under various IPI attacks. The results demonstrate that \method not only provides robust security but also significantly excels in maintaining task utility, outperforming all existing baselines.

\textbf{Competitive Security.} 
In terms of security, \method achieves an average ASR of 0.4\%, which is comparable to the performance of the \textit{LLM-Detector} (0.2\% average ASR). While the latter serves as a strong upper-bound baseline, \method attains a similar level of protection through a much more lightweight probing mechanism. This demonstrates that capturing latent traces is a highly effective alternative to using a dedicated external guardrail model for identifying sophisticated attacks.

\textbf{Preservation of Task Utility.} 
The most distinct advantage of \method is its ability to maintain agent utility under attack. While fine-tuning based defenses~(LLM Detectors) often secure the system by rejecting inputs, which inevitably terminates the task. But our Mitigating Rectifier recovers a substantial portion of the model's functional capability. Specifically, \method provides an average 10\% improvement in UA compared to other defense baselines. In the \textsc{Qwen-3} experiments, for instance, our method maintains a UA of 53.8\%, whereas the next best defense drops below 49\%.

\textbf{Baselines Comparison.} 
Compared to the tool-filtering method \textit{MELON}, which acts as a post-filter for action selection, \method reduces the ASR by an additional 3\% while providing a 10\% UA boost. Since MELON cannot re-run the reasoning process after a filter trigger, it inevitably loses utility. Furthermore, while template-based methods (\textit{Repeat Prompt} and \textit{Delimiting}) show some effectiveness against traditional \textit{Ignore Instruction} attacks, they struggle significantly with adaptive \textit{TrojanTools}. Interestingly, we observe that template methods slightly improve UA on weaker backbones like \textsc{LLaMA} and \textsc{Mistral}, likely because structured prompts help these models better parse user intent despite their limited native reasoning capabilities.

\subsubsection{Efficiency Analysis.} 
We further evaluate the computational overhead of \method compared to representative baselines in Table~\ref{tab:efficiency}. While fine tuning-based defenses like \textsc{Qwen3Guard} offer robust safety, they necessitate extensive fine-tuning (over 10 hours) on massive datasets (1.19M samples), incurring significant resource costs. In contrast, \method is highly computationally agile, requiring less than 2 minutes of training on only 255 samples, with a minimal parameter count of approximately 31k. While heuristic-based defenses (e.g., MELON, Repeat Prompt) incur zero training cost, they rely on rigid, static policies that often fail against adaptive attacks. In summary, \method thus occupies a \textit{\textbf{sweet spot}}: it maintains the near-zero deployment overhead characteristic of heuristic methods while providing the learned, latent-aware security typically associated with heavy-weight fine-tuning.

\subsection{Out-of-Distribution Performance} 
To evaluate the generalization capability of \method, we test a prober trained exclusively on the \textit{TrojanTools} dataset against unseen benchmarks, including \textit{AgentDojo}~\cite{agentdojo} and \textit{InjecAgent}~\citep{zhan2024injecagent}. As presented in Table~\ref{tab:detection_defense_rate}, the results demonstrate that \method generalizes effectively to different distributions without requiring additional fine-tuning. 

Specifically, \method achieves a detection rate exceeding 97\% on \textit{AgentDojo} and remains robust on \textit{InjecAgent} with a detection rate over 80\%. More importantly, the utility recovery rate consistently exceeds 60\% for both \textsc{Qwen} and \textsc{LLaMA} backbones on Qwen3-8B. This strong zero shot transfer performance suggests that \method captures intrinsic latent traces of IPI attacks rather than over-fitting to specific semantic patterns in the training data, confirming its practical viability for real-world, evolving adversarial environments.

\begin{figure}
    \centering
    \includegraphics[width=0.99\linewidth]{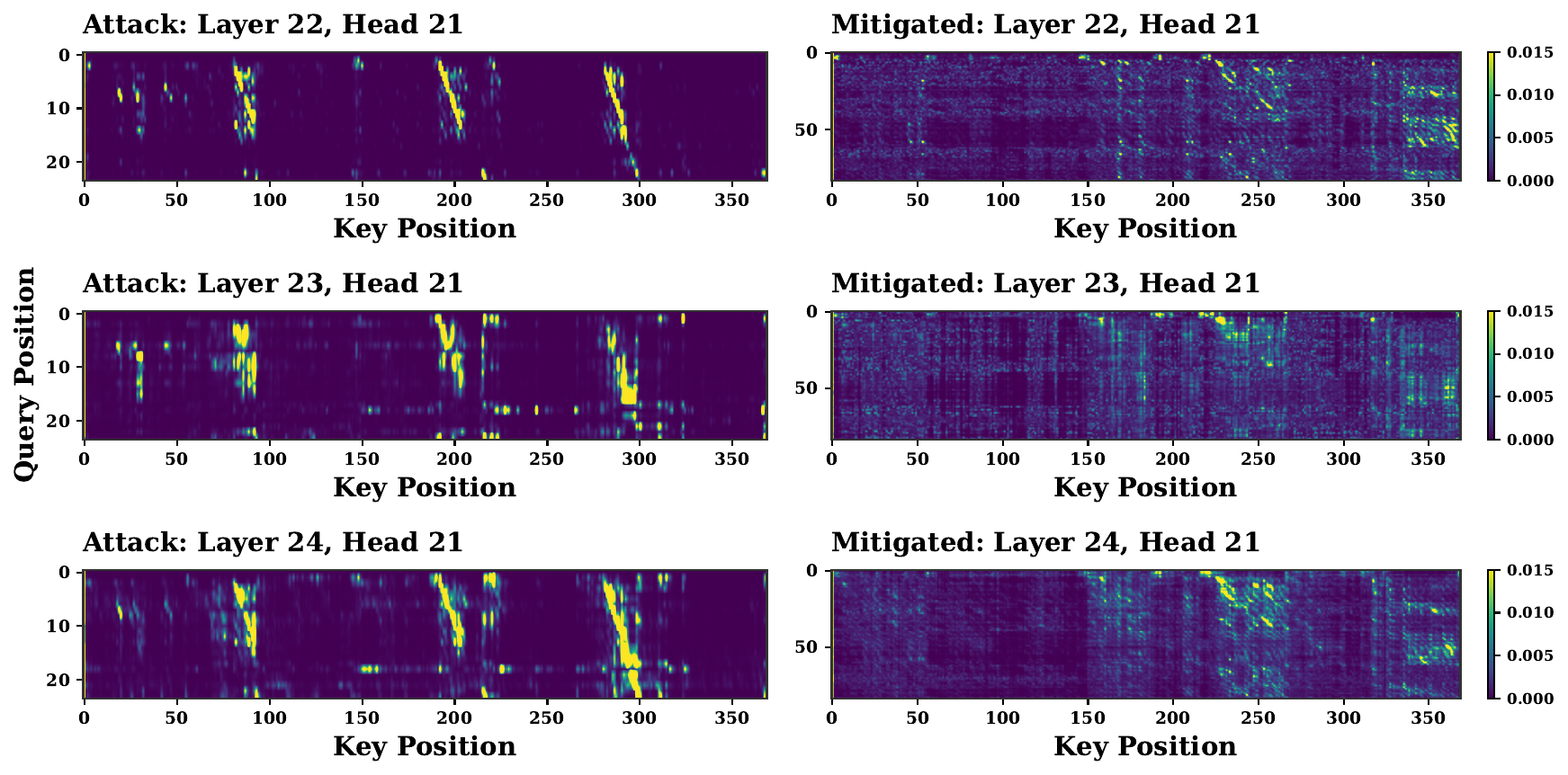}
    \caption{Visualization of Mitigate Rectifier's Functionality}
    \label{fig:visual_attention}
\end{figure}

\begin{table}[t]
\centering
\small
\setlength{\tabcolsep}{4pt}
\renewcommand{\arraystretch}{1.2}
\caption{Effectiveness of \method on Various Agentic MLLMs.}
\resizebox{\columnwidth}{!}{%
\begin{tabular}{l|l|cccccccc}
\toprule
\multirow{2}{*}{\textbf{FMs}} & \multirow{2}{*}{\textbf{BU}} & \multicolumn{2}{c|}{\textbf{No Def.}} & \multicolumn{2}{c|}{\textbf{Repeat}} & \multicolumn{2}{c|}{\textbf{LLM-Det.}} & \multicolumn{2}{c}{\textbf{Ours}} \\
\cmidrule(lr){3-4} \cmidrule(lr){5-6} \cmidrule(lr){7-8} \cmidrule(lr){9-10}
& & \textbf{ASR$\uparrow$} & \textbf{UA$\downarrow$} & \textbf{ASR$\uparrow$} & \textbf{UA$\downarrow$} & \textbf{ASR$\uparrow$} & \textbf{UA$\downarrow$} & \textbf{ASR$\uparrow$} & \textbf{UA$\downarrow$} \\
\midrule
\textsc{Qwen3VL} & 70.5 & 69.9 & 21.3 & 47.8 & 46.5 & 0.0 & 21.9 & 3.8 & 44.9 \\
\textsc{InternVL} & 64.5 & 47.4 & 33.9 & 41.6 & 26.5 & 0.0 & 34.1 & 2.6 & 47.8\\
\textsc{MiniCPM} & 72.0 & 59.7 & 29.0 & 59.6 & 22.0 & 0.0 & 30.3 & 2.4 & 48.8 \\
\midrule
\textbf{Avg} & 69.0 & 59.0 & 28.1 & 49.7 & \second{31.7} & \first{0.0} & 28.8 & \second{2.9} & \first{47.2} \\
\bottomrule
\end{tabular}}
\begin{flushleft}
\scriptsize
\textbf{Notes:} LLM-Detector = Gemini.
\end{flushleft}
\label{tab:MLLMs_performance}
\vspace{-2em}
\end{table}

\subsection{Visualization of Attention Steering}
To provide an intuitive understanding of the Rectifier's mechanism, we visualize the attention maps of a critical head 21 from layer 22-25 in the \textsc{Qwen-3-8B} model. As illustrated in Figure~\ref{fig:visual_attention}, the impact of the \method intervention is evident in the redistribution of attention weights. Under a successful IPI attack (left), the model exhibits an extreme attention collapse, where the focus is disproportionately concentrated on specific token indices (approximately at positions 100 and 300) corresponding to the malicious payload. This localized ``over-focusing'' indicates that the adversarial instruction has hijacked the head's functional priority. Conversely, after applying the MR (right), these anomalous adversarial peaks are effectively suppressed through our selective masking and steering operation. The attention is subsequently dispersed across the broader agent trajectory, restoring the model's ability to integrate global contextual information. This visualization confirms that the Rectifier successfully ``steers'' the latent representation back to the benign task direction, thereby enabling the agent to execute its original plan despite the presence of adversarial triggers. 

%Moreover, due to space limitations, we provide the hyperparameter analysis in Appendix.\ref{hypermeter_analysis}, which presents the best combination of $\gamma$ and $\tau$ is 0.45 and 0.4.

\subsection{Robustness Across Modalities}
To further validate the architectural agnosticism and cross-modal generalizability of \method, we extend our evaluation to MLLMs. Unlike text-only scenarios, multi-modal IPI attacks can be embedded within visual inputs, posing a more complex threat to the agent's latent space. As presented in Table~\ref{tab:MLLMs_performance}, \method demonstrates robust defensive resilience in these vision-language contexts. In terms of security, our method achieves a competitive average ASR of 2.9\%, a drastic reduction compared to the \textit{Repeat Prompt} baseline (49.7\%). While the commercial \textit{LLM-Detector} achieves a zero ASR, its average utility (UA) is restricted to 28.8\%, suggesting a tendency toward binary refusal that disrupts task continuity. In contrast, \method effectively maintains the task execution loop, reaching a 47.2\% UA, which is an absolute improvement of nearly 20\% over Gemini. This underscores that our attention steering mechanism successfully identifies and neutralizes adversarial triggers within multi-modal structures without sacrificing the agent's capabilities.

\section{Conclusion}
This paper introduces ICON, a defense framework against prompt injection attacks—particularly indirect prompt injection. ICON comprises two core defense modules and an adaptive module. Specifically, it employs a Focus Intensity Score to identify critical layers and attention heads, then leverages a latent-space trace prober to precisely detect adaptive attacks. A mitigating rectifier subsequently steers the LLM away from malicious intent toward its intended task. Extensive experiments across multiple LLMs and modalities demonstrate that ICON achieves an exceptional balance among security, utility, and efficiency: it maintains an ASR of just 0.4\% while delivering over 50\% higher utility than commercial-grade baselines. Most notably, its extreme efficiency requiring less than two minutes of training—makes ICON a practical and scalable safeguard for real-world autonomous agents.

\newpage

\section*{Impact Statement}
This paper presents work whose goal is to advance the field of machine learning, specifically in the area of AI security and agent robustness. By developing more resilient autonomous agents, our work helps mitigate the risks associated with adversarial manipulation in large language models. There are many potential societal consequences of our work, most of which contribute to the safe and ethical deployment of AI, and none of which we feel must be specifically highlighted here beyond its contribution to general system reliability.

\bibliography{icml2026}
\bibliographystyle{icml2026}

%%%%%%%%%%%%%%%%%%%%%%%%%%%%%%%%%%%%%%%%%%%%%%%%%%%%%%%%%%%%%%%%%%%%%%%%%%%%%%%
%%%%%%%%%%%%%%%%%%%%%%%%%%%%%%%%%%%%%%%%%%%%%%%%%%%%%%%%%%%%%%%%%%%%%%%%%%%%%%%
% APPENDIX
%%%%%%%%%%%%%%%%%%%%%%%%%%%%%%%%%%%%%%%%%%%%%%%%%%%%%%%%%%%%%%%%%%%%%%%%%%%%%%%
%%%%%%%%%%%%%%%%%%%%%%%%%%%%%%%%%%%%%%%%%%%%%%%%%%%%%%%%%%%%%%%%%%%%%%%%%%%%%%%
% \newpage
% \appendix
% \onecolumn
% \input{appendix}

\end{document}